\documentclass[11pt,a4paper]{article}
\usepackage[hyperref]{acl2019}
\usepackage{times}
\usepackage{latexsym}
\usepackage{graphicx}
\usepackage{amssymb}
\usepackage{caption}
\usepackage{subcaption}
\usepackage{booktabs}
\usepackage{multirow}

\usepackage{url}

\aclfinalcopy % Uncomment this line for the final submission
\title{Towards a Metric for Automated Conversational Dialogue System Evaluation and Improvement}

\author{Jan Deriu \\
  Zurich University of Applied Sciences \\
  \texttt{deri@zhaw.ch} \\\And
  Mark Cieliebak \\
  Zurich University of Applied Sciences \\
  \texttt{ciel@zhaw.ch} \\}

\date{}

\begin{document}
\maketitle
\begin{abstract}
We present "AutoJudge", an automated evaluation method for conversational dialogue systems. The method works by first generating dialogues based on self-talk, i.e. 
dialogue systems talking to itself. Then, it uses human ratings on these dialogues to train an automated judgement model. 
Our experiments show that AutoJudge correlates well with the human ratings and can be used to automatically evaluate dialogue systems, even in deployed systems. 
In a second part, we attempt to apply AutoJudge to improve existing systems. This works well for re-ranking a set of candidate utterances. However, our experiments show that AutoJudge cannot
be applied as reward for reinforcement learning, although the metric can distinguish good from bad dialogues. 
We discuss potential reasons, but state here already that this is still an open question for further research.
\end{abstract}

% !TEX root = acl2019.tex
\section{Introduction}
Conversational dialogue systems (also referred to as chatbots, social bots, or non-task-oriented dialogue systems) allow for a natural conversation between computer and humans. Research on these dialogue systems has recently reemerged due to the availability of large dialogue corpora, \cite{serban2015surveyJournal} as well as the popularization of deep learning \cite{sordoni2015contect,vinyals2015neural,serban2016generative}. %The combination of large amounts of data and deep learning allows for training conversational dialogue systems in an end-to-end fashion where all subsystems of a traditional dialogue system (i.e. natural language understanding, state tracking, dialogue management and natural language generation) are part of the same end-to-end neural network. 

%However, unlike task-oriented systems, conversational dialogue systems, do not have a clearly defined task, whose success can be measured. For task-oriented the task success as well as the dialogue efficiency (i.e. number of turns until success) are used to automatically evaluate and improve the systems. Most notably user satisfaction modelling techniques \cite{Walker:1997:PFE:979617.979652,SCHMITT201512} or user simulations\cite{schatzmann:2006:us_survey,kreyssig2018neural} are used for this.
One major challenge in developing high-quality dialogue systems is the evaluation process. Ideally, an evaluation method should be automated, have a high correlation to human judgements and be able to discriminate between different dialogue strategies. Most common techniques to evaluate conversational dialogue systems rely on crowdsourcing, where human judges are asked to rate the \emph{appropriateness} (or \emph{quality}) of a generated response given a context. Although this procedure allows to discriminate between different strategies, it has several drawbacks: it is time and cost intensive, it has to be redone for every change in dialogue strategy, and the results cannot be used to improve the system.

On the other hand, the automated evaluation is usually performed by applying word-overlap metrics borrowed from the machine translation or text summarization community, which have been shown to correlate poorly to human judgements on the utterance level \cite{liu2016eval}. %Furthermore, these techniques require a gold-standard to compare to, which is usually not  available in deployed systems. 

\paragraph{Trained Metrics.} Recently, the notion of \emph{trained metrics} was introduced for conversational dialogue systems \cite{Lowe2017AutoTuring}. The main idea is that humans rate the generated response of a dialogue system in relation to a given context (i.e. the dialogue history). Based on these ratings, a regression model is trained which models the human judges. For this, the context, the candidate response, and the gold-standard response are used as input and the judgement is predicted. This approach correlates well with human judgements on the turn level as well as on the system level.

However, these metrics rely on a gold-standard and work on static contexts, which is problematic for two reasons. First, as the context is written by humans it does not reflect the behaviour of the dialogue system. Second, it cannot be used in deployed systems where no gold-standard is available. Dynamic context evaluation \cite{Gandhe2016}, on the other hand, usually requires human-computer interaction, which is costly, and puts an additional cognitive strain on the users if they are to rate live during the conversation \cite{SCHMITT201512}. %In this work, we build upon this idea and extend it in the following ways.
%\paragraph{Static vs. Dynamic Context Evaluation.}
%The previously introduced methods rely on a gold standard and a static context. This is problematic for two reasons. First, the context does not reflect the behaviour of the dialogue system. Second, it cannot be used in deployed systems. On the other hand, dynamic context evaluation \cite{Gandhe2016} requires a generated context, which usually requires humans interacting with the dialogue system, which is very costly. In this work, we build upon this idea and extend it in the following ways.
%The previously introduced methods rely on having access to a gold standard, which means that the context is static. This is problematic for two reasons. First, the context is often written entirely by humans, thus, it does not reflect the behaviour of the dialogue system at hand. Second, it restricts the usage of the method, i.e. it cannot be used in deployed systems. %For instance, \cite{Lowe2017AutoTuring} works on a static context, which is taken form a test corpus and the dialogue system is evaluated based on the response it generates for this particularDynamic context evaluation \cite{Gandhe2016}, on the other hand, requires the context to be generated, which usually requires humans interacting with the dialogue system. Furthermore, if the annotations are to be done on the turn level, the users need to rate the dialogue after each turn, which puts a high cognitive demand on the user \cite{SCHMITT201512}. In this work, we build upon this idea and extend it in the following ways.

\paragraph{Contribution.} In this work we propose to automatically generate the dialogues relying on \emph{self-talk}, which is derived from \emph{AlphaGo} self-play \cite{silver2016mastering}. Dialogues are generated by two instances of the same system conversing with each other. Then the automatically generated dialogues are rated by human judges. That is, the judges read the dialogues and rate it on the turn level. Based on these ratings, we train a regression model which learns to predict the ratings of the human judges. Our results show that this method, which we refer to as \emph{AutoJudge}, achieves high correlation to human judgements. Thus, it can be applied to fully automatically assess the quality of a dialogue system without being dependent on gold standard responses.

\paragraph{Applications.} Since our approach is fully automatic and requires no humans in the loop, we want to go one step further and apply it to {\em improve} the dialogue system at hand. More precisely we attempt to apply the metric in two different ways: (i) response ranking similar to \cite{shalyminov-etal-2018-neural,hancock2019learning}, and (ii) reward for reinforcement learning. It turns out that only the re-ranking shows promising results, whereas the metric is not useful as a reward function. This is very surprising, since the trained metric correlates
well to human judgements, and it can discriminate between good and bad utterances. Why this happens, and how it can be resolved,  is  an open research question, which we discuss towards the end of this paper. 

%\input{02_related_work}
%!TEX root = acl2019.tex
\section{Experimental Setup}
Our experimental pipeline follows three phases. First, the data generation phase, where we let the dialogue systems generate dialogues automatically. Second, the data annotation phase, where we rely on crowdsouring to rate the dialogues on the turn level. Third, the improvement phase, where we train an automated judgement model on the annotated data and apply this model to improve the dialogue system. 

\subsection{Dialogue Systems}
For our experiments we relied on the following state-of-the-art dialogue systems (the training details are in Appendix \ref{sec:app_a}):

\paragraph{Seq2Seq.} The Sequence-to-Sequence model as proposed by \cite{vinyals2015neural} consists of an encoder and a decoder. Both modules are based on Long Short-Term Memory cells (LSTM) \cite{hochreiter1997}, where the encoder consumes the last utterance and produces a hidden representation, which is passed as initial state to the decoder to condition the generation process. 
\paragraph{HRED.} The Hierarchical Recurrent Encoder-Decoder (HRED) model proposed by \cite{Serban2016Hier} enhances the Seq2Seq model by a hierarchical encoding procedure. Here, the context-turns are encoded by first encoding each turn separately and then by applying a recurrent encoder over the hidden states of the turns. The decoding procedure is conditioned on the hidden state produced by the context encoder. 
\paragraph{VHRED.} The Hierarchical Latent Variable Encoder-Dcoder model (VHRED) \cite{serban2017latentdialog} enhances the aforementioned HRED model by introducing a stochastic latent variable at the utterance level. This stochastic variable aims to inject variability at the utterance level, which in turn increases the variety of responses a model generates. 
\paragraph{MrRNN.} The Multi-resolution Recurrent Neural Ntwork (MrRNN) \cite{serban2017mrrnn} enhances the HRED model by introducing an abstraction layer. More precisely, the dialogue is modelled by processing the inputs and outputs at various level of abstractions (e.g. at the level of meaning bearing words and the usual word-level). 
\paragraph{DE.} The Dual Encoder (DE) \cite{lowe-EtAl:2015:W15-46} is a selection based model, which differs from the generation based approaches of the aforementioned models. The DE encodes both the context and a candidate response (using the same encoder as the VHRED model) and then classifies if the candidate is a valid response to the given context. 

%For all models, we used a bidirectional LSTM to encode the turns, and a unidirectional LSTM for both the context encoder and decoder. We specify the number of units for the LSTMs to 500, 1000, 1000 for the turn-encoder, context-encoder and decoder respectively. We use the pretrained 300 dimensional FastText embeddings \cite{mikolov2018advances}, which we refine during the training. In order to avoid too large vocabularies, we limit the vocabulary size to 20k distinct tokens. The generative models are trained to minimize the reconstruction error. For the VHRED and MrRNN, we refer to the original papers for the loss function formulation. The Dual Encoder is trained to minimize a contrastive loss function. 

\subsection{Turn-Level Annotation}
We apply \emph{self-talk} to automatically generate dialogues. For this, we sample 100 different contexts randomly from a set of unseen contexts and let the dialogue system generate a dialogue starting from this context, which consist of 10 turns each. For the annotation process, we use Amazon Mechanical Turk (AMT) \footnote{https://www.mturk.com/ } and follow the procedure outlined by \cite{Lowe2017AutoTuring}, i.e. the judges rated the \emph{overall quality} of each turn on a scale from 1 (low quality) to 5 (high quality). Each turn is annotated by three different judges. We required the AMT workers to be from an english speaking country (USA, UK, Ireland or Australia) in order to ensure that they are native speakers, since the generated messages are highly colloquial and make heavy usage of slang. For each annotation, we paid 15 cents, where we assumed that each annotation takes between 60 to 90 seconds. For the selection of the final turn-label, we apply the MACE procedure \cite{hovy2013mace}, which learns confidence scores for the annotators. Our final dataset consists of a total of 500 annotated dialogues, which amounts to 5000 annotated pairs of contexts and responses. 

\subsection{AutoJudge}
Similarly to the \emph{ADEM} procedure proposed by \cite{Lowe2017AutoTuring}, we train a regression model on the annotated data. For this, we use the pre-trained context and response encoder from the VHRED model. Unlike \emph{ADEM}, our dialogues are generated automatically, thus, we do not have access to a gold-standard response. For this reason, we use the following scoring function:
$score(c, r) = (c^{T}Mr - \alpha)/\beta$
where $M \in \mathbb{R}^{d \times d}$ is a learned similarity matrix, $\alpha , \beta$ are scalar constants, and $c, r$ are the context and response embeddings respectively. The model is optimized to minimize the mean squared error between the predicted ratings and the human judgements. 
%We optimize \emph{AutoJudge} using the \emph{Adam} optimizer \cite{kingma2014adam}.

\subsection{Improving Dialogue Systems}
%The main goal of our \emph{AutoJudge} is to find applications, where it can improve the dialogue systems.
Since \emph{AutoJudge} is fully automated, we apply it to improve the existing dialogue systems. For this, we implemented the following two applications: as reward for reinforcement learning (RL), and as re-ranking candidate utterances. 
\paragraph{Re-Ranking.}
Given a list of responses from the five aforementioned dialogue systems for a given context, \emph{AutoJudge} re-ranks them by their predicted score. In our experiments, we use the dialogue systems, which we trained for the self-talk experiment, i.e. we re-rank the outputs of the five aforementioned dialogue systems. Thus, the re-ranker serves as a meta-selection module.
%\paragraph{Filter}
%For the training set filtering, we rate all the context-response pairs in the training set. That is, we use the ratings to eliminate all the training samples, which have rating below 3. 
%The second, technique uses the ratings as a sample-weight. That is, the ratings are projected into the range between 0 and 1 and then the loss of each sample is weighted with the transformed rating. The second technique is a softer version of the first one. 
\paragraph{Reinforcement Learning Reward.}
We apply the predicted ratings as reward in the RL framework. For this, we apply the Policy Gradient formulation, as done in  \cite{D16-1127}, which is defined as follows:
$\bigtriangledown  J_{RL}(\theta) = \sum_{i} \bigtriangledown \log p(\textit{r}_i | \textit{c}_i) \times \sum_i  R(\textit{r}_i, \textit{c}_i)$
, where $r_i$ and $c_i$ are the response and context in the $i^{th}$ turn, $R(\textit{r}_i, \textit{c}_i)$ is the predicted reward by \emph{AutoJudge}, and $\sum_{i} \log p(\textit{r}_i | \textit{c}_i)$ is the reconstruction error.

% !TEX root = acl2019.tex
\section{Results and Discussion}
In our experiments we use the Twitter Dialogue Corpus \cite{Ritter:2011:DRG:2145432.2145500}\footnote{We use the IDs provided by \cite{serban2017latentdialog}, which can be found here: www.iulianserban.com/Files/TweetIDs.zip}. The Twitter Dialogue Corpus provides social interactions, which we believe to be a good basis for being annotated via crowdsouring. 

\paragraph{Data Aggregation.}
The turn-level ratings provide us with 5000 annotated pairs of context and responses. The distribution over the labels is balanced (i.e. each class is represented between $19\%$ and $21\%$ of the cases). However, the agreement scores among the human judges is rather low: the median pairwise Spearman correlation between two judges is only at $0.403$. Furthermore, the MACE procedure reports on the confidence score (between 0 and 1) of single judges, which is used as basis for selecting the final label. The average confidence is at only $0.15$. We assume that these problems stem from the high degree of subjectivity of the problem. 
%For task-oriented systems this problem can be avoided by using a clearly defined guidelines \cite{SCHMITT201512}. However, it is not clear how to precisely define \emph{quality} in open-domain chit-chat dialogues. 
%make point on correlation when trained on 4 systems and test on 5th

%% figure
%%----------------------------------------------------------------------------
%\begin{figure}
%	\begin{center}
%        \begin{tabular}{@{}c@{\hspace{5mm}}c@{\hspace{5mm}}c@{}}
%		\includegraphics[width=0.22\textwidth]{figures/DataDistr0_cropped.pdf} &
%		\includegraphics[width=0.22\textwidth]{figures/DataDistr1_cropped.pdf} \\	
%	    \end{tabular}
%	\end{center}
%	\caption{Data distributions for both the overall data and for the systems.}
%\label{fig:data_distr}
%\end{figure}
%%----------------------------------------------------------------------------

\paragraph{AutoJudge.}
\begin{table}[h!]
\vspace{-1mm}
\begin{center}
\begin{small}
\resizebox{.4\textwidth}{!}{
\begin{tabular}{l|cccc} 
\hline
%\abovespace
\textsc{} & Pearson Corr& Spearman's Rho & MAE \\
\hline 
\textsc{Convo Split}&	 0.573 & 0.577 & 0.928 \\
\textsc{System Split}&	 0.544 & 0.53 & 0.984 \\
\hline
\end{tabular}
}
\end{small}
\\~\\
  \vspace{1mm}
\end{center}
\caption{Average correlations between the judgements predicted by AutoJudge and the human judgement. 
\textsc{Convo Split} denotes the cross-validation split according to the contexts and  \textsc{System Split} denotes the cross-validation split according to the dialogue system.
}
\label{table:correlations}
\end{table}

We train \emph{AutoJuge} using k-fold cross validation. There are two ways of splitting the data into folds, in order to ensure that all turns of the same dialogue are in the same fold. First, we group the 100 contexts into 10 folds, thus, each fold consists of 50 dialogues (i.e. 10 contexts times the number of dialogue systems), this is denoted as \textsc{Convo Split}. The second option is to split the data according to the system which created the conversation, which evaluates the performance of \emph{AutoJuge} in rating dialogues of unseen dialogue systems. We denote this as  \textsc{System Split}.
In Table \ref{table:correlations}, we report the average Pearson correlation, Spearman's rho and mean absolute error (MAE) over all folds for the \emph{conversation split} and the \emph{system split}. With moderate correlations of $0.573$ on the dialogue level, we get results which are comparable to \cite{Lowe2017AutoTuring}, where ADEM achieves a Pearson correlation of $0.436$. Note that we cannot directly compare our results to BLEU score and ADEM, since these base their predictions on gold standards, which we do not have in our setting. An interesting result is the \emph{System Split}, i.e. that our approach is able to maintain a high correlation ($0.544$) with the ratings of a dialogue system when removing the data of that system from the training, which is not the case in \cite{Lowe2017AutoTuring} where the correlation for a different system dropped significantly.

%Lastly, we evaluate the correlation on the system level. For this, we compute the average ratings over all turns from each dialogue system. We compare the average ratings from the judges with the average predicted ratings over 4000 randomly sampled dialogues. Table \ref{table:system_correlations} shows these ratings. Although the absolute scores do not correspond between the humans and \emph{AutoJudge}, the Pearson correlation is at $0.793$. Thus, \emph{AutoJudge} is a viable metric for discriminating between different dialogue system strategies. 
 
%
%
%
%\begin{table}[h]
%\vspace{-1mm}
%\begin{center}
%\begin{small}
%\resizebox{.3\textwidth}{!}{
%\begin{tabular}{l|cc} 
%\hline
%%\abovespace
%&\multicolumn{2}{c}{Avg. Turn Ratings} \\
%\hline
%Systems & Human& Predicted \\
%\hline 
%\textsc{Seq2Seq}&	 			3.512 & 3.699 \\
%\textsc{HRED}&	 				2.956 & 3.638 \\
%\textsc{VHRED}&	 				3.392 & 3.676 \\
%\textsc{MrRNN}&	 				2.792 & 3.626 \\
%\textsc{Dual Encoder}&	 		2.612 & 3.378 \\
%\hline
%\end{tabular}
%}
%\end{small}
%\\~\\
%  \vspace{1mm}
%\end{center}
%\caption{Human ratings compared to the predicted ratings on the system level. One rating is the average rating over all turns of one dialogue system. The Pearson correlation between the avgerage human rating and the average predicted rating is $0.793$.}
%\label{table:system_correlations}
%\end{table}

\paragraph{Answer Selection.} 
In order to evaluate the improvements achieved by the re-ranking method, we sample a disjoint set of 100 new contexts and apply \emph{self-talk} to generate conversations. Then, we use AMT to let humans judge the automatically generated conversations on the dialogue level (i.e. a rating for the entire dialogue as opposed to turn-based ratings). We compare the performance of the five base dialogue systems to the performance of the re-ranking strategy. 
Table \ref{table:applications} shows the average scores for each dialogue system. 
Our results show that the \emph{re-ranking} approach works very well. It raises the score to $3.47$, which is $0.16$ points higher than the best base-system (i.e. \textsc{Seq2Seq}). 

\begin{table}[h]
\vspace{-1mm}
\begin{center}
\begin{small}
\resizebox{.3\textwidth}{!}{
\begin{tabular}{l|cc} 
\hline
%\abovespace
Systems & Dialogue Level Rating  \\
\hline 
\textsc{Seq2Seq}&	 			\textit{3.31}  \\
\textsc{HRED}&	 				2.78  \\
\textsc{VHRED}&	 				3.20  \\
\textsc{MrRNN}&	 				2.37 \\
\textsc{Dual Encoder}&	 		2.02 \\
\hline \hline
%\textsc{HRED-Filter}&	 		2.52 \\
\textsc{Re-Ranking}&	 	\textbf{3.47} \\
\end{tabular}
}
\end{small}
\\~\\
  \vspace{0mm}
\end{center}
\caption{Human judgements on the dialogue level for each dialogue system. For this, a each dialogue system (the five base-systems and the re-ranking system) generate 100 dialogues using self-talk, which human judges rated on the dialogue level. Here we see the average ratings for each system.}
\label{table:applications}
\end{table}

\paragraph{Reinforcement Learning.} When we apply \emph{AutoJudge} as reward resulted in suboptimal dialogues. Although the return increases over time (from $21.74$ to $37.41$ over 80 episodes), the dialogues which the policy generates are often incoherent or completely useless. This seems counter-intuitive when taking into account the aforementioned high correlation scores. We believe that the main reason for the suboptimal behaviour is that \emph{AutoJudge} does not have enough coverage during training. Thus, very bad responses (e.g. empty responses, repeating responses, convergence to a single universal response) tend to receive high scores, since the training data for \emph{AutoJudge} does not include these kinds of responses. However, it is not clear how to stabilize \emph{AutoJudge} to handle these cases. For instance, by artificially enhancing the training data for \emph{AutoJudge} with negative examples, the Pearson correlation score drops to $0.50$ without any impact on the reinforcement learning. %Thus, it is still an open question on how to apply trained metrics as reward to reinforcement learning. 
%There needs to be a second signal, which measures whether the responses are sensible or not. This could be achieved by means of a language model, which rates the perplexity of a sentence or by means of a generative-adversarial network \cite{Goodfellow2014GAN}. 

%% figure
%%----------------------------------------------------------------------------
%\begin{figure}
%	\begin{center}
%		\includegraphics[width=0.22\textwidth]{figures/rolling_avg.pdf}
%	\end{center}
%	\caption{Data distributions for both the overall data and for the systems.}
%\label{fig:data_distr}
%\end{figure}
%%----------------------------------------------------------------------------

% !TEX root = acl2019.tex
\section{Conclusion}
Our results show that \emph{AutoJudge} correlates well to human judgements and it is useful to measure the progress of a dialogue system, as it is able to discriminate among different strategies. Furthermore, it generalizes well to unseen strategies for the same domain. Since \emph{AutoJudge} is independent of a gold-standard it can be applied to deployed systems where gold-standards are not available. Finally, it shows promising results when applied as answer selection module. As a next step, we intend to apply \emph{AutoJudge} onto human-computer dialogues to measure the viability of \emph{AutoJudge} in a real-world setting. 

In this work we tried to use \emph{AutoJudge} as a reward for reinforcement learning, which resulted in suboptimal dialogues. The main reason seems to be that \emph{AutoJudge} cannot properly handle the bad utterance that are generated during the initial phase of reinforcement learning. This is surprising, since \emph{AutoJudge} is able to distinguish good and bad  utterances of fully-trained systems. This seems to indicate that there are different types of "bad" utterances, and we need to adapt the training mechanism of \emph{AutoJudge} if we want to apply it not only to evaluation, but also to  improving dialogue systems. Our results indicate that trained metrics suffer from instabilities, which might be caused by the size of the dataset.

One major issue is that it is not clear which aspects \emph{AutoJudge} captures. Although the correlation between the human judgements and the outputs of \emph{AutoJudge} are high, we cannot make any statement about what aspects of the context or the response are relevant for the predicted rating. This is a fundamental problem with the evaluation of conversational dialogue systems, as there is no clear definition for "adequate" responses. Thus, an important future work problem is the investigation into the definition of "adequacy" for conversational dialogue systems.

We conjecture that this might apply also to other automated metrics, thus, this is an important research question that needs to be addressed if we want to understand how to better train and optimize dialogue systems.

%Lastly, we show the limitations of \emph{AutoJudge}, that is, it cannot be used as a reward for reinforcement learning. However, it shows promising results when applied as answer selection module. We believe that future work on how to apply trained metrics to reinforcement learning should consider ways to regularize the procedure using language models to assess if a response is reasonable or not. 

\section{Acknowledgements}
This paper has been partially funded by the LIHLITH project, supported by ERA-NET CHIST-ERA and Swiss National Science Foundation (20CH21\_174237).

\bibliography{acl2019}
\bibliographystyle{acl_natbib}

\appendix
% !TEX root = acl2019.tex

\clearpage

\begin{table*}[h!] 
 \centering 
 \footnotesize 
 \resizebox{0.9\textwidth}{!}{\begin{minipage}{1\textwidth}
 \begin{tabular}{|l|l|l|l|l|} 
 \hline 
 \textbf{\#} & \textbf{\#Turn} & \textbf{Utterance} & \textbf{Pred.} & \textbf{Human}\\
 \hline 
 \multirow{1}{*}{\begin{minipage}{0.01\textwidth} 1 \end{minipage}} 
 &\multirow{2}{*}{\begin{minipage}{0.01\textwidth} 1 \end{minipage}} & 
\multirow{2}{*}{\begin{minipage}{0.70\textwidth} @user ur prolly tired now , arent u ? wanna sleep , don't cha ? (: \end{minipage}} & 
\multirow{2}{*}{\begin{minipage}{0.01\textwidth}       4.67 \end{minipage}} & 
\multirow{2}{*}{\begin{minipage}{0.01\textwidth}       5.00 \end{minipage}} 
\\ & & & & \\ 
  \cline{2-5} 
 &\multirow{2}{*}{\begin{minipage}{0.01\textwidth} 2 \end{minipage}} & 
\multirow{2}{*}{\begin{minipage}{0.70\textwidth} @user atm jus chillin nd chattin nd listening to music :P haha xxx \end{minipage}} & 
\multirow{2}{*}{\begin{minipage}{0.01\textwidth}       3.70 \end{minipage}} & 
\multirow{2}{*}{\begin{minipage}{0.01\textwidth}       2.00 \end{minipage}} 
\\ & & & & \\ 
  \cline{2-5} 
 &\multirow{2}{*}{\begin{minipage}{0.01\textwidth} 3 \end{minipage}} & 
\multirow{2}{*}{\begin{minipage}{0.70\textwidth} @user x factor tickets came so im siva happy dancing :) might do some fanfic xx wbu ? xx \end{minipage}} & 
\multirow{2}{*}{\begin{minipage}{0.01\textwidth}       3.26 \end{minipage}} & 
\multirow{2}{*}{\begin{minipage}{0.01\textwidth}       1.00 \end{minipage}} 
\\ & & & & \\ 
  \cline{2-5} 
 &\multirow{2}{*}{\begin{minipage}{0.01\textwidth} 4 \end{minipage}} & 
\multirow{2}{*}{\begin{minipage}{0.70\textwidth} @user goood :) do u want to be in my 1d fanbook ? xx \end{minipage}} & 
\multirow{2}{*}{\begin{minipage}{0.01\textwidth}       3.07 \end{minipage}} & 
\multirow{2}{*}{\begin{minipage}{0.01\textwidth}       2.00 \end{minipage}} 
\\ & & & & \\ 
  \cline{2-5} 
 &\multirow{2}{*}{\begin{minipage}{0.01\textwidth} 5 \end{minipage}} & 
\multirow{2}{*}{\begin{minipage}{0.70\textwidth} @user : o gosh ppl these days :/ ... ok tell me everything 2moz p . s send me the link so i can send it to et :) xx dont stress it :) \end{minipage}} & 
\multirow{2}{*}{\begin{minipage}{0.01\textwidth}       2.89 \end{minipage}} & 
\multirow{2}{*}{\begin{minipage}{0.01\textwidth}       1.00 \end{minipage}} 
\\ & & & & \\ 
  \cline{2-5} 
 &\multirow{2}{*}{\begin{minipage}{0.01\textwidth} 6 \end{minipage}} & 
\multirow{2}{*}{\begin{minipage}{0.70\textwidth} @user yer im fine thanks xxx \end{minipage}} & 
\multirow{2}{*}{\begin{minipage}{0.01\textwidth}       3.18 \end{minipage}} & 
\multirow{2}{*}{\begin{minipage}{0.01\textwidth}       3.00 \end{minipage}} 
\\ & & & & \\ 
  \cline{2-5} 
 &\multirow{2}{*}{\begin{minipage}{0.01\textwidth} 7 \end{minipage}} & 
\multirow{2}{*}{\begin{minipage}{0.70\textwidth} @user okay , thanks xxx \end{minipage}} & 
\multirow{2}{*}{\begin{minipage}{0.01\textwidth}       3.42 \end{minipage}} & 
\multirow{2}{*}{\begin{minipage}{0.01\textwidth}       1.00 \end{minipage}} 
\\ & & & & \\ 
  \cline{2-5} 
 &\multirow{2}{*}{\begin{minipage}{0.01\textwidth} 8 \end{minipage}} & 
\multirow{2}{*}{\begin{minipage}{0.70\textwidth} @user :) you okay ? xxx \end{minipage}} & 
\multirow{2}{*}{\begin{minipage}{0.01\textwidth}       3.26 \end{minipage}} & 
\multirow{2}{*}{\begin{minipage}{0.01\textwidth}       1.00 \end{minipage}} 
\\ & & & & \\ 
  \cline{2-5} 
 &\multirow{2}{*}{\begin{minipage}{0.01\textwidth} 9 \end{minipage}} & 
\multirow{2}{*}{\begin{minipage}{0.70\textwidth} @user yeah im good thanks :) you ? xx \end{minipage}} & 
\multirow{2}{*}{\begin{minipage}{0.01\textwidth}       3.11 \end{minipage}} & 
\multirow{2}{*}{\begin{minipage}{0.01\textwidth}       4.00 \end{minipage}} 
\\ & & & & \\ 
 \hline 
 \hline
 \multirow{1}{*}{\begin{minipage}{0.01\textwidth} 2 \end{minipage}} & 
\multirow{2}{*}{\begin{minipage}{0.01\textwidth} 0 \end{minipage}} & 
\multirow{2}{*}{\begin{minipage}{0.70\textwidth} if i could i would hurt you ... a lot ! ! i'm so sick of you blaming me for everything that goes wrong ! ! \end{minipage}} & 
\multirow{2}{*}{\begin{minipage}{0.01\textwidth}       4.86 \end{minipage}} & 
\multirow{2}{*}{\begin{minipage}{0.01\textwidth}       5.00 \end{minipage}} 
\\ & & & & \\ 
  \cline{2-5} 
 &\multirow{2}{*}{\begin{minipage}{0.01\textwidth} 1 \end{minipage}} & 
\multirow{2}{*}{\begin{minipage}{0.70\textwidth} @user awh ... is everything ok sweetie ? \end{minipage}} & 
\multirow{2}{*}{\begin{minipage}{0.01\textwidth}       4.72 \end{minipage}} & 
\multirow{2}{*}{\begin{minipage}{0.01\textwidth}       5.00 \end{minipage}} 
\\ & & & & \\ 
  \cline{2-5} 
 &\multirow{2}{*}{\begin{minipage}{0.01\textwidth} 2 \end{minipage}} & 
\multirow{2}{*}{\begin{minipage}{0.70\textwidth} @user thanks ... it is ... i just can't wait to see how it is ... \end{minipage}} & 
\multirow{2}{*}{\begin{minipage}{0.01\textwidth}       3.72 \end{minipage}} & 
\multirow{2}{*}{\begin{minipage}{0.01\textwidth}       3.00 \end{minipage}} 
\\ & & & & \\ 
  \cline{2-5} 
 &\multirow{2}{*}{\begin{minipage}{0.01\textwidth} 3 \end{minipage}} & 
\multirow{2}{*}{\begin{minipage}{0.70\textwidth} @user wait , what ? \end{minipage}} & 
\multirow{2}{*}{\begin{minipage}{0.01\textwidth}       3.51 \end{minipage}} & 
\multirow{2}{*}{\begin{minipage}{0.01\textwidth}       5.00 \end{minipage}} 
\\ & & & & \\ 
  \cline{2-5} 
 &\multirow{2}{*}{\begin{minipage}{0.01\textwidth} 4 \end{minipage}} & 
\multirow{2}{*}{\begin{minipage}{0.70\textwidth} @user i have a new phone . i have to get it . \end{minipage}} & 
\multirow{2}{*}{\begin{minipage}{0.01\textwidth}       2.54 \end{minipage}} & 
\multirow{2}{*}{\begin{minipage}{0.01\textwidth}       3.00 \end{minipage}} 
\\ & & & & \\ 
  \cline{2-5} 
 &\multirow{2}{*}{\begin{minipage}{0.01\textwidth} 5 \end{minipage}} & 
\multirow{2}{*}{\begin{minipage}{0.70\textwidth} @user how much do you pay ? \end{minipage}} & 
\multirow{2}{*}{\begin{minipage}{0.01\textwidth}       2.20 \end{minipage}} & 
\multirow{2}{*}{\begin{minipage}{0.01\textwidth}       3.00 \end{minipage}} 
\\ & & & & \\ 
  \cline{2-5} 
 &\multirow{2}{*}{\begin{minipage}{0.01\textwidth} 6 \end{minipage}} & 
\multirow{2}{*}{\begin{minipage}{0.70\textwidth} @user pay attention \end{minipage}} & 
\multirow{2}{*}{\begin{minipage}{0.01\textwidth}       2.65 \end{minipage}} & 
\multirow{2}{*}{\begin{minipage}{0.01\textwidth}       1.00 \end{minipage}} 
\\ & & & & \\ 
  \cline{2-5} 
 &\multirow{2}{*}{\begin{minipage}{0.01\textwidth} 7 \end{minipage}} & 
\multirow{2}{*}{\begin{minipage}{0.70\textwidth} @user i would have to do that \end{minipage}} & 
\multirow{2}{*}{\begin{minipage}{0.01\textwidth}       1.82 \end{minipage}} & 
\multirow{2}{*}{\begin{minipage}{0.01\textwidth}       1.00 \end{minipage}} 
\\ & & & & \\ 
  \cline{2-5} 
 &\multirow{2}{*}{\begin{minipage}{0.01\textwidth} 8 \end{minipage}} & 
\multirow{2}{*}{\begin{minipage}{0.70\textwidth} @user you could of been in my head . \end{minipage}} & 
\multirow{2}{*}{\begin{minipage}{0.01\textwidth}       2.27 \end{minipage}} & 
\multirow{2}{*}{\begin{minipage}{0.01\textwidth}       1.00 \end{minipage}} 
\\ & & & & \\ 
  \cline{2-5} 
 &\multirow{2}{*}{\begin{minipage}{0.01\textwidth} 9 \end{minipage}} & 
\multirow{2}{*}{\begin{minipage}{0.70\textwidth} @user yeah , i don't know . \end{minipage}} & 
\multirow{2}{*}{\begin{minipage}{0.01\textwidth}       3.03 \end{minipage}} & 
\multirow{2}{*}{\begin{minipage}{0.01\textwidth}       1.00 \end{minipage}} 
\\ & & & & \\ 
 \hline 
 \hline
 \multirow{1}{*}{\begin{minipage}{0.01\textwidth} 3 \end{minipage}} & 
\multirow{2}{*}{\begin{minipage}{0.01\textwidth} 0 \end{minipage}} & 
\multirow{2}{*}{\begin{minipage}{0.70\textwidth} \# hashtag @user @user @user @user @user @user @user @user \end{minipage}} & 
\multirow{2}{*}{\begin{minipage}{0.01\textwidth}       4.47 \end{minipage}} & 
\multirow{2}{*}{\begin{minipage}{0.01\textwidth}       5.00 \end{minipage}} 
\\ & & & & \\ 
  \cline{2-5} 
 &\multirow{2}{*}{\begin{minipage}{0.01\textwidth} 1 \end{minipage}} & 
\multirow{2}{*}{\begin{minipage}{0.70\textwidth} @user thanks for the follow . do you have at bat on your phone yet ? \end{minipage}} & 
\multirow{2}{*}{\begin{minipage}{0.01\textwidth}       4.95 \end{minipage}} & 
\multirow{2}{*}{\begin{minipage}{0.01\textwidth}       5.00 \end{minipage}} 
\\ & & & & \\ 
  \cline{2-5} 
 &\multirow{2}{*}{\begin{minipage}{0.01\textwidth} 2 \end{minipage}} & 
\multirow{2}{*}{\begin{minipage}{0.70\textwidth} @user there's a lot going on ... but i think is good to say what u think here and at the boards too . specially there ! \end{minipage}} & 
\multirow{2}{*}{\begin{minipage}{0.01\textwidth}       3.55 \end{minipage}} & 
\multirow{2}{*}{\begin{minipage}{0.01\textwidth}       2.00 \end{minipage}} 
\\ & & & & \\ 
  \cline{2-5} 
 &\multirow{2}{*}{\begin{minipage}{0.01\textwidth} 3 \end{minipage}} & 
\multirow{2}{*}{\begin{minipage}{0.70\textwidth} @user pretty much just as well as the current macbook pros do ... \end{minipage}} & 
\multirow{2}{*}{\begin{minipage}{0.01\textwidth}       2.42 \end{minipage}} & 
\multirow{2}{*}{\begin{minipage}{0.01\textwidth}       1.00 \end{minipage}} 
\\ & & & & \\ 
  \cline{2-5} 
 &\multirow{2}{*}{\begin{minipage}{0.01\textwidth} 4 \end{minipage}} & 
\multirow{2}{*}{\begin{minipage}{0.70\textwidth} @user how well does it run photoshop / illustrator ? \end{minipage}} & 
\multirow{2}{*}{\begin{minipage}{0.01\textwidth}       2.67 \end{minipage}} & 
\multirow{2}{*}{\begin{minipage}{0.01\textwidth}       2.00 \end{minipage}} 
\\ & & & & \\ 
  \cline{2-5} 
 &\multirow{2}{*}{\begin{minipage}{0.01\textwidth} 5 \end{minipage}} & 
\multirow{2}{*}{\begin{minipage}{0.70\textwidth} @user this is like sixth sense . can i have your games consoles ? \end{minipage}} & 
\multirow{2}{*}{\begin{minipage}{0.01\textwidth}       3.00 \end{minipage}} & 
\multirow{2}{*}{\begin{minipage}{0.01\textwidth}       1.00 \end{minipage}} 
\\ & & & & \\ 
  \cline{2-5} 
 &\multirow{2}{*}{\begin{minipage}{0.01\textwidth} 6 \end{minipage}} & 
\multirow{2}{*}{\begin{minipage}{0.70\textwidth} @user probably , but i don't know if it'd be appropriate in this case . i try only to use our funds for things i know i need to know . \end{minipage}} & 
\multirow{2}{*}{\begin{minipage}{0.01\textwidth}       2.15 \end{minipage}} & 
\multirow{2}{*}{\begin{minipage}{0.01\textwidth}       1.00 \end{minipage}} 
\\ & & & & \\ 
  \cline{2-5} 
 &\multirow{2}{*}{\begin{minipage}{0.01\textwidth} 7 \end{minipage}} & 
\multirow{2}{*}{\begin{minipage}{0.70\textwidth} @user well we can scan thru a few joints i'm workin on , or take the traditional route n pen one down \end{minipage}} & 
\multirow{2}{*}{\begin{minipage}{0.01\textwidth}       2.09 \end{minipage}} & 
\multirow{2}{*}{\begin{minipage}{0.01\textwidth}       1.00 \end{minipage}} 
\\ & & & & \\ 
  \cline{2-5} 
 &\multirow{2}{*}{\begin{minipage}{0.01\textwidth} 8 \end{minipage}} & 
\multirow{2}{*}{\begin{minipage}{0.70\textwidth} @user i learned gordon's rhythm patterning in methods classes , not much orff \end{minipage}} & 
\multirow{2}{*}{\begin{minipage}{0.01\textwidth}       2.21 \end{minipage}} & 
\multirow{2}{*}{\begin{minipage}{0.01\textwidth}       1.00 \end{minipage}} 
\\ & & & & \\ 
  \cline{2-5} 
 &\multirow{2}{*}{\begin{minipage}{0.01\textwidth} 9 \end{minipage}} & 
\multirow{2}{*}{\begin{minipage}{0.70\textwidth} @user its like a long story with sequence of events lmao \end{minipage}} & 
\multirow{2}{*}{\begin{minipage}{0.01\textwidth}       1.76 \end{minipage}} & 
\multirow{2}{*}{\begin{minipage}{0.01\textwidth}       1.00 \end{minipage}} 
\\ & & & & \\ 
\hline
 \end{tabular} 
 \caption{Randomly sampled output. The conversation is sampled at random and \emph{AutoJudge} rates each turn.} 
 \end{minipage}}
 \label{tbl:app_examples_base1} 
 \end{table*}

\clearpage

\section{Training Details}
\label{sec:app_a}
\paragraph{Model Training.} For all models, we used a bidirectional LSTM to encode the turns, and a unidirectional LSTM for both the context encoder and decoder. We specify the number of units for the LSTMs to 500, 1000, 1000 for the turn-encoder, context-encoder and decoder respectively. We use the pretrained 300 dimensional FastText embeddings \cite{mikolov2018advances}, which we refine during the training. In order to avoid too large vocabularies, we limit the vocabulary size to 20k distinct tokens. The generative models are trained to minimize the reconstruction error. For the VHRED and MrRNN, we refer to the original papers for the loss function formulation. The Dual Encoder is trained to minimize a contrastive loss function $log \sigma(c^{T}r_{Ture}) + \sum_{n \in N} log \sigma(-c^{T}r_{n})$, where c is the context encoding, $r_{True}$ is the correct response encoding and N is a set of negative samples. For each training sample we sampled 10 negative examples uniformly at random from the training set. All models are optimized using the \emph{Adam} optimizer \cite{kingma2014adam}, with a $lr=0.001$ and a batch size of 80. 

\paragraph{AutoJudge Training}
We trained \emph{AutoJudge} using the pre-trained VHRED model to encode the context and the response. During the training only the matrix M gets optimized. We also experimented with non-linear transformation on these encodings, which did not yield any improvements. Similar to \cite{Lowe2017AutoTuring}, we use $\alpha = 0.01$ and $\beta=32$. \emph{AutoJudge} is optimized using \emph{Adam} optimizer \cite{kingma2014adam}, with a $lr=0.001$ and a batch size of 512.

%\section{Data Distribution Plot}
% figure
%----------------------------------------------------------------------------
%\begin{figure}
%	\begin{center}
%        \begin{tabular}{@{}c@{\hspace{5mm}}c@{\hspace{5mm}}c@{}}
%		\includegraphics[width=0.22\textwidth]{figures/DataDistr0_cropped.pdf} &
%		\includegraphics[width=0.22\textwidth]{figures/DataDistr1_cropped.pdf} \\	
%	    \end{tabular}
%	\end{center}
%	\caption{Plots for the data distribution. On the left side, we see the distribution for each label. On the right side, we see the distribution for for each system.}
%\label{fig:data_distr}
%\end{figure}
%----------------------------------------------------------------------------

\section{Reinforcement Learning}
For reinforcement learning, we use the pre-trained HRED system as our initial policy. We apply Policy Gradient as described above. We experimented with various episode batch sizes  (1, 10, 100, 1000), i.e. in sample n episodes at once to reduce variance. However, it had no impact on the performance. We also experimented with different formulations, i.e. using Advantage Actor Critics in order to reduce the variance. 

\begin{figure}
	\begin{center}
		\includegraphics[width=0.45\textwidth]{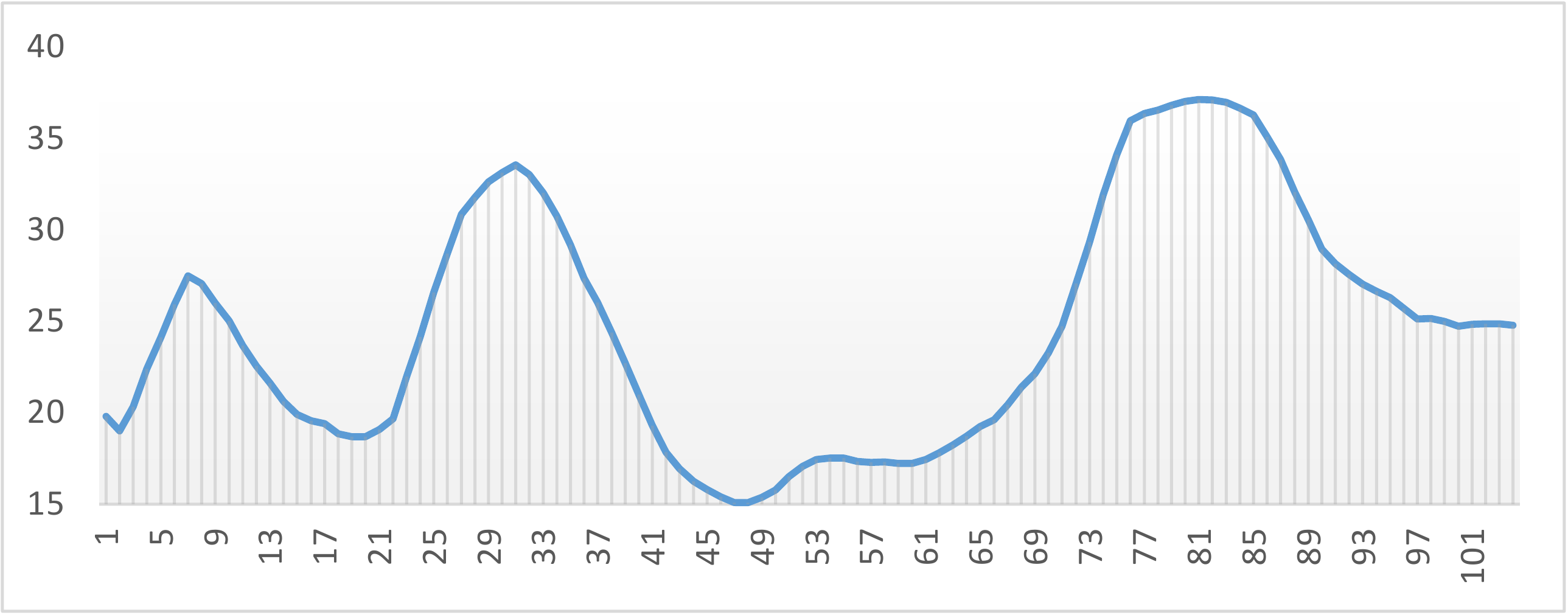}
	\end{center}
	\caption{Data distributions for both the overall data and for the systems.}
\label{fig:rolling_avg}
\end{figure}

In Table \ref{fig:rolling_avg}, we show the rolling average return over the course of 100 episodes. We used a batch size of 100 and we used the standard Policy Gradient formulation. The reward oscillates, which is due to finding new local maxima. He maximal observed reward is at $37$ after 80 episodes. However, the generated dialogues are all empty, i.e. the dialogue system always returns the "end-of-sequence" token right away.

% figure
%----------------------------------------------------------------------------
%----------------------------------------------------------------------------

%
%\section{Data Annotation Tool}
% figure
%----------------------------------------------------------------------------
%\begin{figure}
%	\begin{center}
%        \begin{tabular}{@{}c@{\hspace{5mm}}c@{\hspace{5mm}}c@{}}
%		\includegraphics[width=0.42\textwidth]{figures/annotation_tool1.png}\\
%		\includegraphics[width=0.42\textwidth]{figures/annotation_tool2.png} \\	
%	    \end{tabular}
%	\end{center}
%	\caption{Plots for the data distribution. On the left side, we see the distribution for each label. On the right side, we see the distribution for for each system.}
%\label{fig:data_distr}
%\end{figure}
%----------------------------------------------------------------------------

\end{document}